\documentclass[letterpaper, 10 pt, conference]{ieeeconf}  

\usepackage{graphicx}
\usepackage{amsmath}
\usepackage{amssymb}
\usepackage{bbm}
\usepackage{xspace}
\usepackage{tabularx} 
\usepackage{threeparttable}
\usepackage{booktabs}
\usepackage{multirow}
\usepackage{makecell}
\usepackage{subcaption}
\usepackage{booktabs}  
\usepackage{cite}
\usepackage{tabularx}
\usepackage{array} 
\usepackage{pifont}
\usepackage{hyperref}

\newcommand{\invig}{InViG\xspace}
\newcommand{\invigmodel}{\textit{\invig-Base}\xspace}
\newcommand{\figref}[1]{Fig. \ref{#1}}
\newcommand{\secref}[1]{Sect. \ref{#1}}
\newcommand{\equaref}[1]{Eq. \ref{#1}}

\newcommand{\tabref}[1]{Table \ref{#1}}
\newcommand{\image}{I}
\newcommand{\expr}{e}
\newcommand{\q}[1]{q_{#1}}
\newcommand{\ans}[1]{a_{#1}}

\newcommand{\tgt}{x^*}

\newcommand{\xmark}{}%

\IEEEoverridecommandlockouts                              

\title{\LARGE \bf
\invig: Open-Ended Interactive Visual Grounding in Human-Robot Interaction with 500K Dialogues
}

\author{Hanbo Zhang$^{*1}$, Jie Xu$^{*1}$, Yuchen Mo$^{1}$, Tao Kong$^{1}$ 
\thanks{* Equal Contribution. $^1$ ByteDance Research}
}

\begin{document}

\maketitle
\thispagestyle{empty}
\pagestyle{empty}

\begin{abstract}

Ambiguity is ubiquitous in human communication. 
Previous approaches in Human-Robot Interaction (HRI) have often relied on predefined interaction templates, leading to reduced performance in realistic and open-ended scenarios.
To address these issues, we present a large-scale dataset, \invig, for interactive visual grounding under language ambiguity.
Our dataset comprises over 520K images accompanied by open-ended goal-oriented disambiguation dialogues, encompassing millions of object instances and corresponding question-answer pairs.
Leveraging the \invig dataset, we conduct extensive studies and propose a set of baseline solutions for end-to-end interactive visual disambiguation and grounding, achieving a 45.6\% success rate during validation. 
To the best of our knowledge, the \invig dataset is the first large-scale dataset for resolving open-ended interactive visual grounding, presenting a practical yet highly challenging benchmark for ambiguity-aware HRI.
Codes and datasets are available at: \href{https://openivg.github.io}{https://openivg.github.io}.

\end{abstract}

\section{INTRODUCTION}


Robots are gradually becoming more prevalent in our homes. Being an important part of our daily lives, it is crucial for robots to understand visual concepts and comprehend natural language instructions. Additionally, uncertainties and ambiguities are inherent in our daily communication. To effectively handle these uncertainties, robots must actively engage in interactions to gather more information. This process helps them develop more accurate models of the world, leading to better decision-making. Previous research has demonstrated the significance of interaction in reducing failures and improving user experience \cite{tellex2014asking, whitney2017reducing, hatori2018interactively, shridhar2020ingress, zhanglu2021invigorate, yang2022interactive, mo2023towards}. 
However, they often rely on predefined interaction templates, which can lead to confusing or problematic questions and hinder their ability to handle ambiguous language instructions in challenging scenarios.

\begin{figure}[t]
 \center{\includegraphics[width=0.47\textwidth]{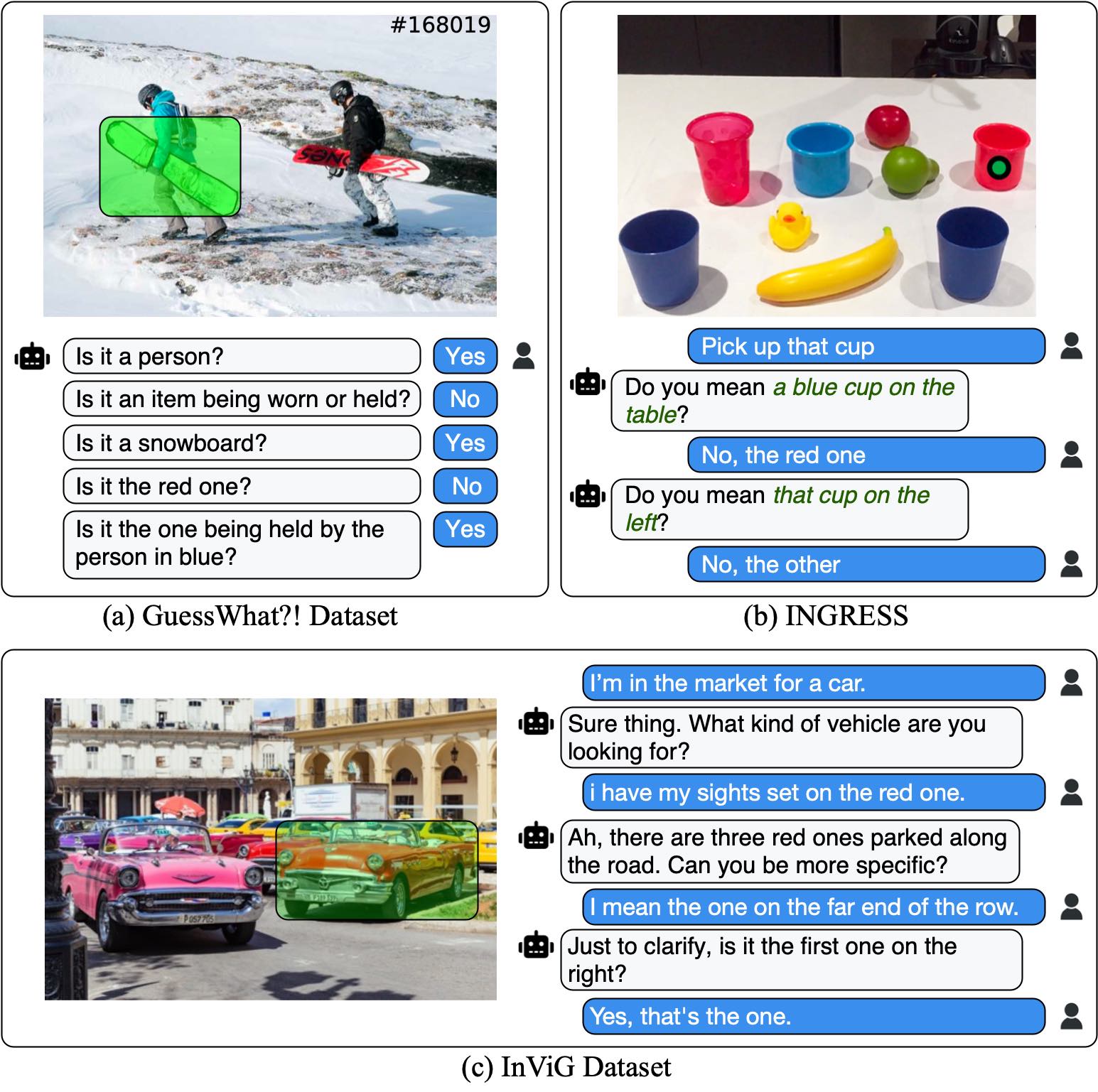}}
 \caption{\invig 500K dataset compared to previous works. (a) GuessWhat?! \cite{de2017guesswhat} dataset, which only contains questions with answers from ``Yes'', ``No'', and ``N/A''; (b) INGRESS \cite{shridhar2020ingress}, which constrains the questions to follow predefined templates; (c) Our \invig 500K with free-form instructions, questions, and answers, designed to facilitate open-ended goal-oriented dialogues for human-robot disambiguation.}
 \label{fig:introduction}
 \vspace{-15pt}
\end{figure}

Recent advancements in large-scale visual-language models (VLMs)~\cite{lu2019vilbert, su2019vl, lu202012, chen2020uniter, zhang2021vinvl, cho2021unifying, radford2021learning, zeng2022xvlm, alayracflamingo} have shown promise in modeling visually-grounded HRI using end-to-end neural networks. 
These models, pre-trained on large amounts of image and text data, exhibit potential in understanding natural language \cite{kamath2021mdetr, li2022grounded, ge2022miles}, grounding symbolic concepts in visual inputs \cite{shridhar2022cliport, subramanian2022reclip}, generating open-ended text or verbal responses \cite{hu2022scaling, chen2022visualgpt}, and even reasoning from knowledge \cite{liu2022matcha}. 
Despite these achievements, multi-turn HRI remains challenging, particularly when reasoning over historical information due to the scarcity of visually-grounded interaction data.

In this paper, we introduce the Interactive Visual Grounding dataset (\invig dataset) and benchmarks for this task.
The \invig dataset is designed to train the neural networks to interact directly with users in natural language using raw images and texts as inputs.
For the first part, our dataset contains over 21K human-to-human visually-grounded dialogs collected based on our open-source online chat program.
Furthermore, based on the \invig 21K dataset, we train models to generate 500K human-robot disambiguation dialogues automatically and leverage ChatGPT to enrich the diversity of languages.
Our aim is to enable the model to learn text generation externally from ChatGPT while being trained visually.
Based on the proposed benchmark, we show that our simple yet efficient baseline can archive a 35.6\% success rate in the interactive visual grounding task when trained solely on \invig 21K dataset, and moreover, a 45.6\% success rate after pre-training on \invig 500K dataset with the help of a state-of-the-art object detector \cite{zhou2022detecting}.

Our contributions include three points:
\begin{itemize}
    \item We present the first large-scale dataset including 520K images specifically designed for object-oriented open-ended interactive visual grounding and disambiguation.
    \item We propose a simple yet strong baseline solution to interactive visual grounding and provide a set of promising results.
    \item We conduct extensive empirical studies exploring different formulations, training strategies, and data combinations, providing comprehensive analysis to facilitate further research in interactive visual grounding for HRI.
\end{itemize}

\section{RELATED WORK}

\begin{figure}[t]
 \center{\includegraphics[width=0.47\textwidth]{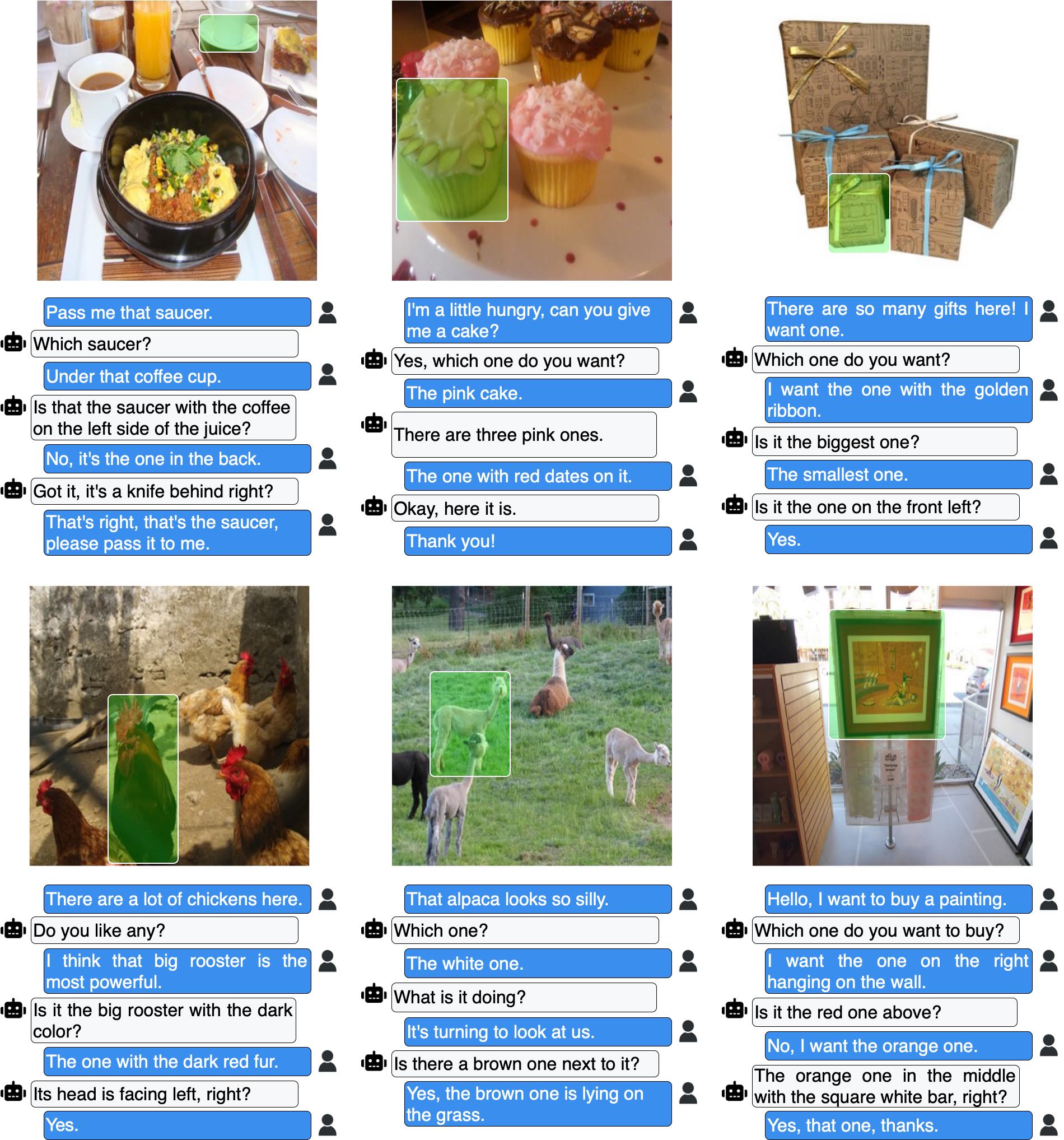}}
 \caption{Examples of \invig 21K dataset.}
 \label{fig:invig21kexamples}
 \vspace{-15pt}
\end{figure}

\paragraph{Disambiguation in HRI} Disambiguation is a practical yet still challenging field in HRI due to the inevitable ambiguity and vagueness of languages.
It has been shown that with a specific task setting and a predefined interaction corpus, verbal instruction generation is typically easier than recognition \cite{thomaz2016computational}.
Therefore, early works usually rely on a predefined corpus to generate verbal expressions for disambiguation to, for example, navigate with the supervision of users~\cite{kruijff2006clarification, hemachandra2015information}, collaborate with humans in complicated tasks~\cite{rosenthal2010effective}, or recover from failures~\cite{deits2013clarifying, tellex2014asking}.
With deep vision methods, it has been demonstrated promising to generate free-form expression with raw images as inputs \cite{johnson2016densecap, yang2017dense, liu2017referring, zheng2022towards}, and hence facilitate the open-ended interaction in open-world scenarios \cite{shridhar2020ingress, zhanglu2021invigorate, yang2022interactive, mo2023towards}.
Instead of following a fixed set of words, the agents may generate object-specific descriptions to make the interaction more natural and flexible.
Yet, such methods are still limited by a set of templates, sometimes making the interaction confusing and even annoying.

\paragraph{Datasets for HRI}
With progress in deep learning, end-to-end approaches have drawn great attention.
As is notoriously known, data-driven approaches are always data-intensive.
Therefore, some works have contributed large-scale datasets for HRI.
Image Chat \cite{shuster2018image} is proposed to train agents for emotional interaction.
The robots are expected to respond differently given different emotional states.
Visual Dialog \cite{das2017visual} is designed following a similar setting of Visual Question Answering (VQA) \cite{antol2015vqa}, but the answer to questions in it may depend on the dialog histories.
GuessWhat?! \cite{de2017guesswhat} is inspired by the well-known interaction game that involves two persons at the same time, one for the Guesser and the other for the Oracle.
The Guesser will be guessing the target of the Oracle by iteratively asking judgment questions, and the Oracle can only answer `Yes' or `No' to complete the game.
Following these datasets, end-to-end approaches have shown great potential for language-based interaction \cite{pang2020visual, tu2021learning, matsumori2021unified}.
More impressively, by training the interaction models end-to-end simply using supervised learning, it even learns to track the dialogue states during interaction \cite{pang2020visual}.
These results show promise to achieve natural HRI through data.
More datasets can be related to visual-language interaction like VQA \cite{antol2015vqa, hudson2019gqa, li2021adversarial, goyal2017making, marino2019ok, gokhale2020vqa}, Visual Reasoning \cite{suhr2018corpus}, Visual Grounding \cite{yu2016modeling}, and Image Captioning \cite{krishna2017visual, chen2015microsoft, sharma2018conceptual}.
Nevertheless, they can hardly be applied to data-driven HRI directly since they only include single-turn interactions.
On the other hand, the most related work GuessWhat?! \cite{de2017guesswhat} simplified the problem too much by restricting the answers to be chosen from ``Yes'' and ``No'', and hence is limited in natural interactions.
In this paper, we contribute a dataset for interactive visual grounding to reduce the gap between current multi-modal models and practical application scenarios in robotics.

\begin{figure}[t]
 \center{\includegraphics[width=0.47\textwidth]{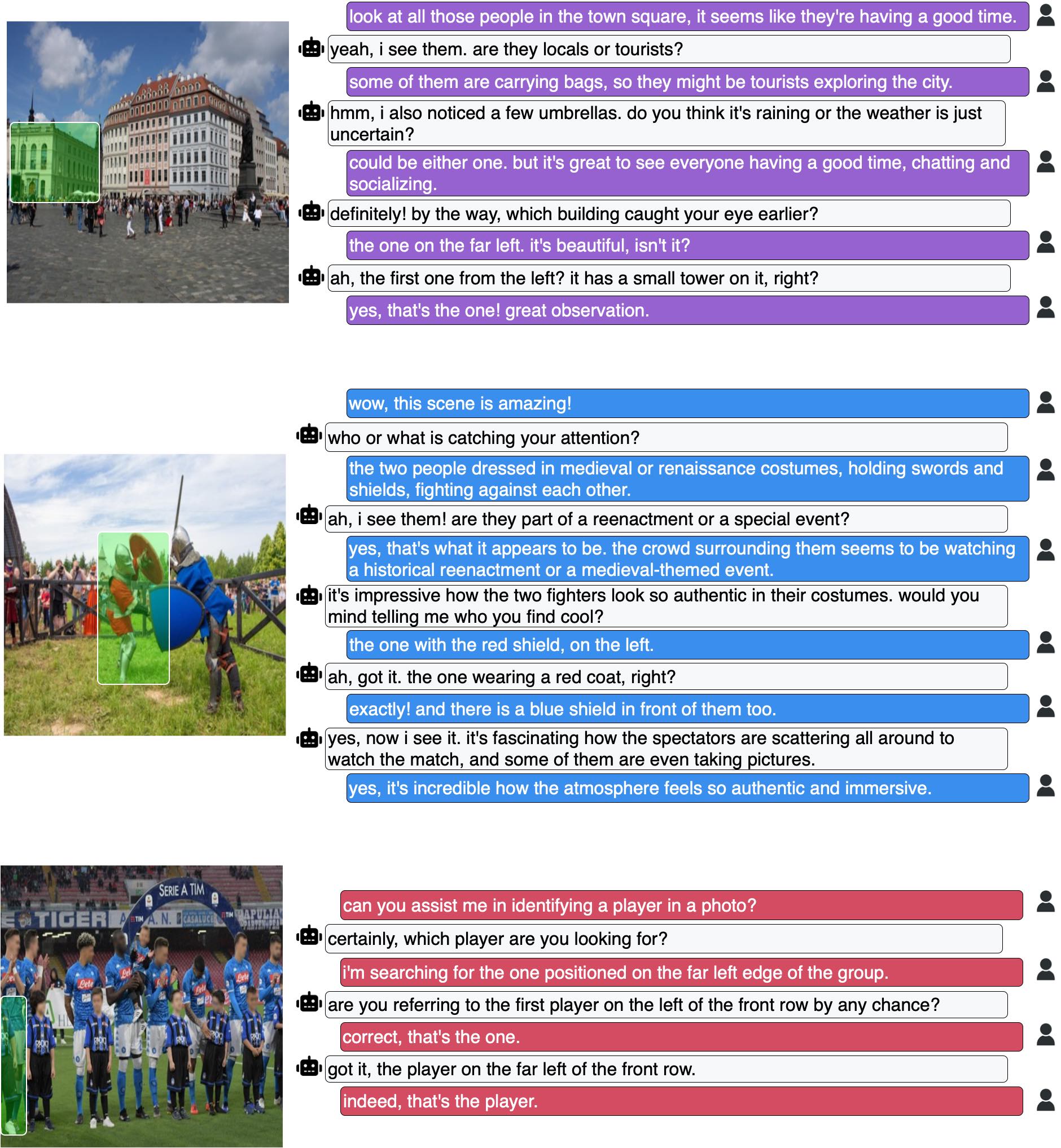}}
 \caption{Examples of \invig 500K dataset.}
 \label{fig:invig500kexamples}
  \vspace{-15pt}
\end{figure}

\begin{table*}[t]  
\caption{\invig Dataset Compared to Other Related Datasets}  
\begin{tabularx}{\textwidth}{>{\raggedright\arraybackslash}p{50pt}*{6}{>{\centering\arraybackslash}X}cccc} 
\toprule  
 & \#Image & \#Dialogue & \#Object & \#Class &\#Question & \#Token & Object-Oriented? & Open-Ended?& Interactive? & Ambiguity? \\
\midrule  
VQA 2.0 & 123K & - & - & - & 658K & 13M & \xmark & \xmark & \xmark & \xmark \\ 
RefCOCO & 20K & - & 197K & 80 & - & 530K & \checkmark & \checkmark & \xmark & \xmark \\
Image Chat & 202K & 202K &  - & - & 401K & 4.7M & \xmark & \checkmark & \checkmark & \xmark \\
Visual Dialog & 125K & 125K & - & - & 1.3M & 11M & \xmark & \checkmark & \checkmark & \xmark \\
GuessWhat?! & 67K & 155K & 1.3M & 80 & 822K & 5.8M & \checkmark & \xmark & \checkmark & \checkmark\\
\midrule
\invig  & \bf 526K & \bf 526K & \bf 43M & \bf $>$500 & \bf 1.9M & \bf 80M &  \checkmark & \checkmark & \checkmark & \checkmark \\
\bottomrule  
\end{tabularx}  
\vspace{-15pt}
\label{tab:data_comp}
\end{table*} 

\paragraph{Deep Learning Models for HRI}
Recently, large-scale language models (LLMs) \cite{brown2020language, zhang2022opt, scao2022bloom, chowdhery2022palm} are shocking the world with their surprisingly emergent zero-shot and few-shot learning abilities \cite{wei2022emergent}.
By simply training the model with intensive text data, such models unite most natural language tasks by text generation.
Though powerful and impressive in natural language tasks, current LLMs still lack efficiency in multi-modal scenes, especially ones that involve the alignment of visual and linguistic concepts.
The most recent works have tried to either assemble or finetune from LLMs to keep the emergent abilities while learning to align visual and linguistic embeddings \cite{alayracflamingo, li2023blip, huang2023language}.
They, nevertheless, are still not capable enough to be deployed in specific task settings like interactive visual grounding due to the lack of data.

\section{\invig Dataset}

\subsection{Overview}

We start with manually annotated interactive dialog collection using crowd-sourcing.
In this phase, we sample and filter 21K images from OpemImages Dataset \cite{kuznetsova2020open}, which consists of massive images with object instances and class-level annotations.
Based on the sampled images, we recruit annotators to label each image with one or more targets and human-to-human dialogues.

With 21K labeled data, we further develop an annotation system to automatically generate HRI data, based on which we further generate 500K goal-oriented disambiguation dialogues
in extremely low costs.
Therefore, in total, our \invig dataset contains more than 520K dialogues for interactive visual grounding.

We demonstrate the comparison between \invig Dataset and previous works in \tabref{tab:data_comp}.
In summary, \invig dataset is proposed to solve the problem of object-oriented open-ended interactive ambiguity in HRI, which widely appears in daily communications between humans.
Therefore, differentiated from all previous works, \invig dataset contains extensive interactive disambiguation data to facilitate the development of HRI systems.

\subsection{Data Collection}
\label{sec:invig21k_collection}


We hope our dataset contains images commonly seen in daily life and vary in ambiguity.
To do so, we sample images from OpenImages dataset~\cite{kuznetsova2020open} with
a two-stage filter.

In the first stage, we remove those images with inappropriate contents, like dense insects and animals like \textit{spiders}, \textit{ladybugs}, \textit{scorpions}, and \textit{marine invertebrates}.
Also, classes that are parts of one another, like \textit{human body}, \textit{human ear}, and \textit{human teeth}, are also filtered out.
Besides, we also filter out most images with low ambiguity, which is defined as the maximum number of objects belonging to the same class.

In the second stage, we propose a Bayesian Filter on the object classes to further identify if an image is suitable for our task.
To be specific, we first randomly sample a small image set and manually classify these images by their suitability into a positive set $I_{pos}$ and a negative set $I_{neg}$. 
For every image $I$ that contains class $c_1, \dots, c_n $, we assume that the appearance of each class $c_i$ ($1\leq i\leq n$) in each image is independent of each other.
Therefore, we have
\begin{align}
\label{eq:1}
    p(I | c_1, \dots, c_n) \propto \prod_{i=1}^{n} p(c_{i} | I)
\end{align}
We approximate $p(c_{i} | I\in I_{pos})$ and $p(c_{i} | I\in I_{neg})$ by simply counting the frequency that each object class appears in positive images and negative images, respectively.
Then, we can apply \equaref{eq:1} to estimate the probability of each image being positive.
Finally, we sort the scores and keep only those images with a positive probability over 0.5, which means they probably portray a good \invig scenario.

Intuitively, the first stage cleans a large portion of undesired images efficiently, while the second stage refines the results by filtering out images that cannot be handled simply using heuristics.

\subsection{Data Annotation}

To collect interaction data, we developed an online interactive website that enables image-based chat between two users.
To start the interaction, each user will be paired randomly with a partner.
During the interaction, each user can choose to skip the current image or reset all labeled data if unexpected situations are encountered.
We force the ``User'' who plays the role of the real user to start the conversation by telling the ``Agent'' what he/she is interested in.
Due to ambiguity in languages, the ``Agent'' needs to ask questions to disambiguate.
We force two annotators to take turn-talking.
Hence, whether a conversation should stop is totally up to the ``Agent''.
That is if the ``Agent'' gets enough information, and finally, locates the targets, it will stop the conversation actively.
During labeling the targets, the ``Agent'' is forced to upload the results first, after which the results will be sent to the ``User'' for checking.

In total, we spent one month recruiting 230 annotators to label the disambiguation dialogues.
Each annotator is required to participate in the label of at most 300 images, considering the diversity of languages with different backgrounds.
To streamline the annotation process and ensure its effectiveness, we utilize specially developed image-based chat tools to collect all the labels.
Finally, we collected more than 21K images and the corresponding targets with disambiguation dialogues.
We have shown some examples of the dataset together with their labels in \figref{fig:invig21kexamples}.

\begin{figure}[t]
    \centering  
      
    \begin{subfigure}[b]{0.22\textwidth}  
        \centering  
        \includegraphics[width=\textwidth]{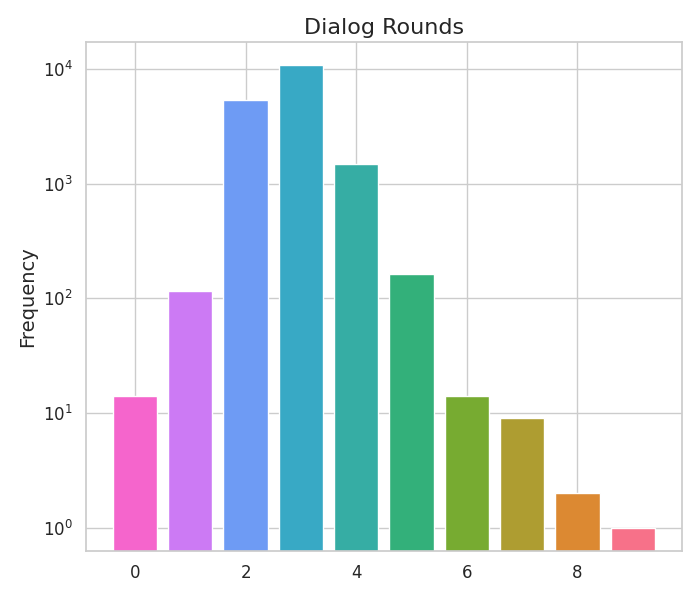}  
        \caption{Dialog rounds}  
        \label{fig:invig21k_dialog_round_dist}  
    \end{subfigure}  
    \hfill  
    \begin{subfigure}[b]{0.25\textwidth}  
        \centering  
        \includegraphics[width=\textwidth]{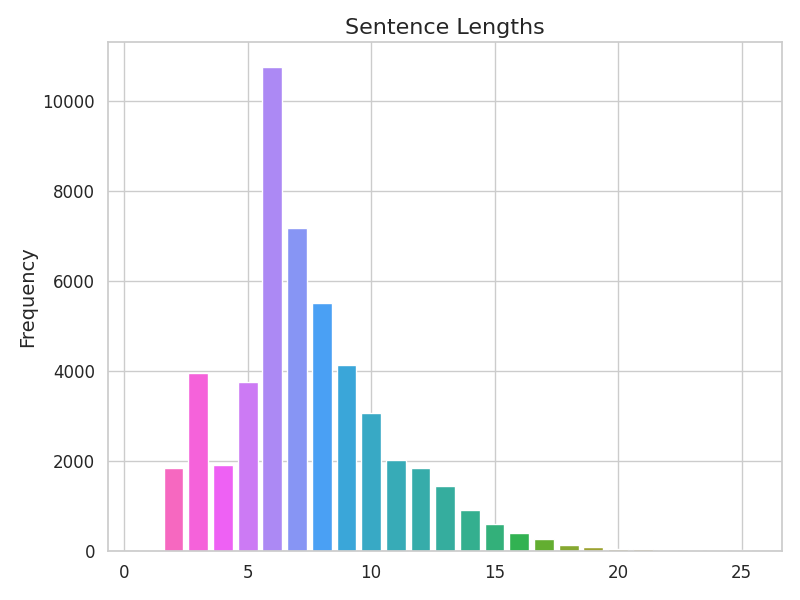}  
        \caption{Sentence lengths}  
        \label{fig:invig21k_sent_len_dist}  
    \end{subfigure}  
      
    \caption{Statistics of dialogues in \invig 21K dataset.}  
    \label{fig:invig21k_dialog_stat}  
\end{figure}  

\begin{figure}[t]
 \center{\includegraphics[width=0.47\textwidth]{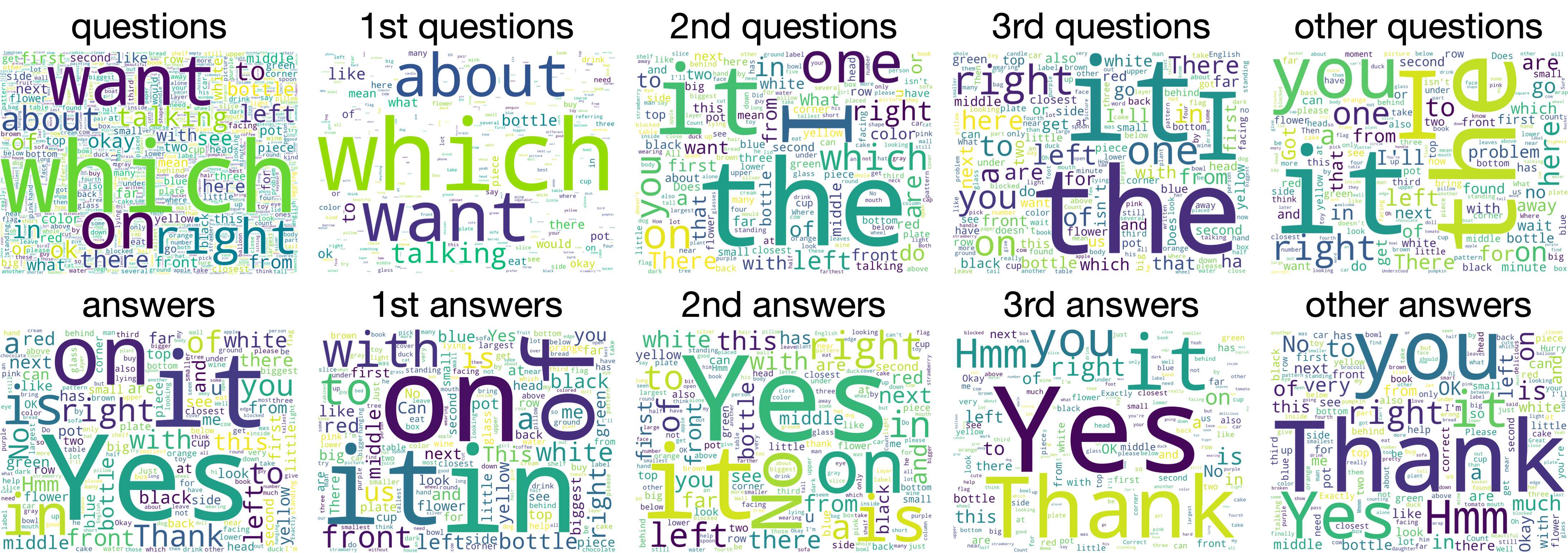}}
 \caption{Word clouds of \invig 21K dataset.}
 \label{fig:invig21k_word_cloud}
 \vspace{-15pt}
\end{figure}

\begin{figure}[t]
    \centering  
      
    \begin{subfigure}[b]{0.22\textwidth}  
        \centering  
        \includegraphics[width=\textwidth]{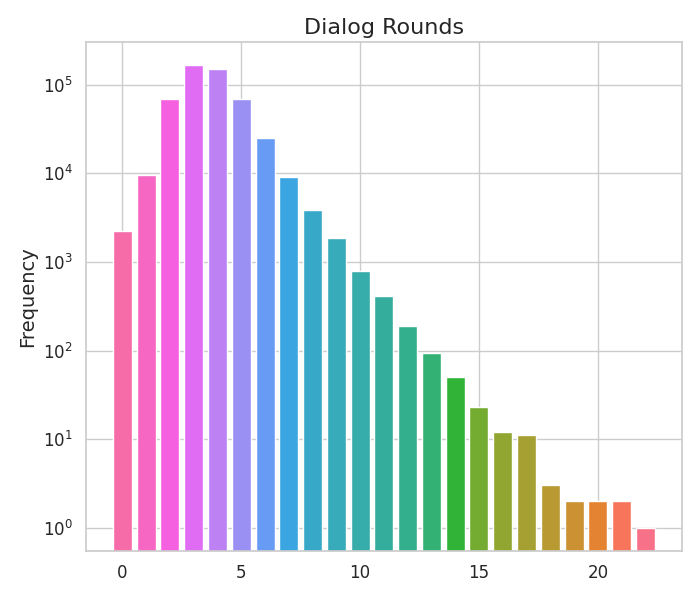}  
        \caption{Dialog rounds}  
        \label{fig:invig500k_dialog_round_dist}  
    \end{subfigure}  
    \hfill  
    \begin{subfigure}[b]{0.25\textwidth}  
        \centering  
        \includegraphics[width=\textwidth]{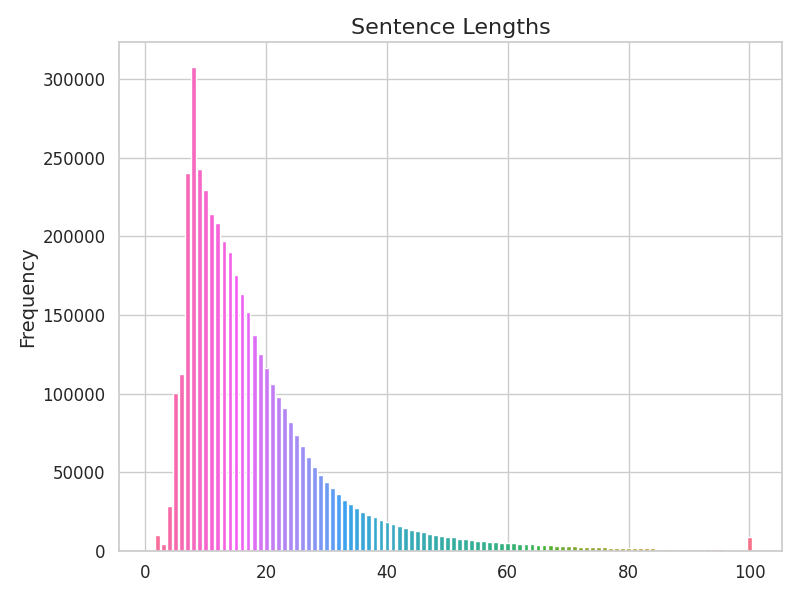}  
        \caption{Sentence lengths}  
        \label{fig:invig500k_sent_len_dist}  
    \end{subfigure}  
      
    \caption{Statistics of dialogues in \invig 500K dataset.}  
    \label{fig:invig500k_dialog_stat}  
\end{figure}  

\begin{figure}[t]
 \center{\includegraphics[width=0.47\textwidth]{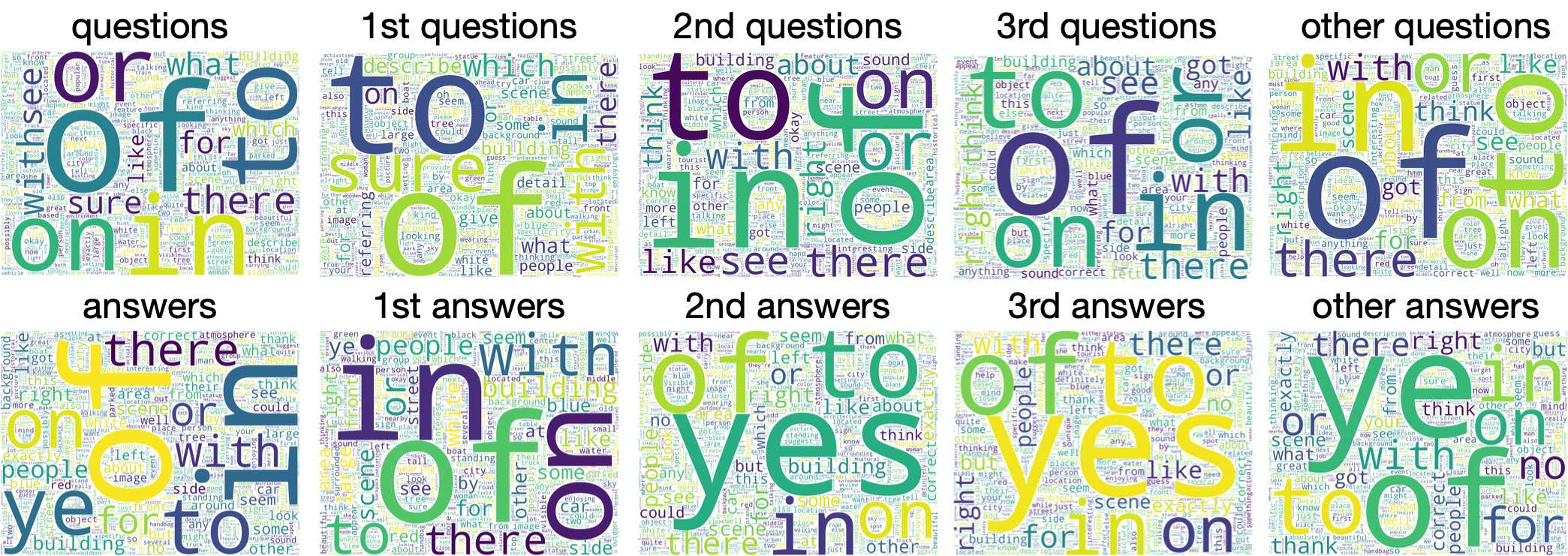}}
 \caption{Word clouds of \invig 500K dataset.}
 \label{fig:invig500k_word_cloud}
  \vspace{-15pt}
\end{figure}

\subsection{From 21K to 520K}

Due to the time and labor-intensive nature of manual annotation, we developed an automatic annotation system to generate disambiguation dialogues only given unlabeled images. 
It enables us to efficiently collect a large-scale dataset with diverse visual and linguistic labels.
By doing so, we collected 500K automatically images and generated dialogues, with diversified visual and linguistic labels.

In this phase, we choose the SAM dataset \cite{kirillov2023segment} as the data source considering the diversity of visual inputs.
Besides, it also contains many more object segments in each image than OpenImages \cite{kuznetsova2020open}, making it too hard for human annotators yet suitable for automatic data generation and collection.
The 500K images are randomly sampled from the Segment Anything dataset.
However, we found that there are around 60\% objects only occupying less than 5\% area in the image, making it too small to be appropriate for our task.
Therefore, we assign the weights for each object according to their occupied areas, meaning that larger objects will be more likely selected.

To label disambiguation dialogues automatically, we train models to interact with each other using the \invig 21K dataset together with other related public datasets \cite{lin2014microsoft, sharma2018conceptual, chen2015microsoft, kazemzadeh2014referitgame, yu2016modeling, zhu2023minigpt, liu2023visual, das2017visual, de2017guesswhat}.
Our models follow the encoder-decoder transformer architectures for visual grounding and text generation \cite{zeng2021multi, zeng2022xvlm, wang2022ofa}.
Concretely, we first train an object captioner to generate a query expression for the selected target, which is the start of each dialogue.
Based on the generated query, a well-trained questioner and oracle will speak alternatively to generate the dialogue until the questioner stops talking.
Finally, a visual grounding model will output the bounding box according to the dialogue and the input image.
To ensure the diversity of language, we then use ChatGPT \cite{openai2023gpt} to rephrase the generated dialogues by our trained models.
Concretely, we prompt ChatGPT to consider the diversity of personality, vocation, and language proficiency during the dialogue augmentation, so that the models can adapt to different kinds of users.
Besides, we also set restrictions in the prompt, to make the output precise and concise. 
Some examples are shown in \figref{fig:invig500kexamples}.
We can see that the generated dialogues contain more diversified expressions compared to the manually labeled ones as shown in \figref{fig:invig21kexamples}.

\subsection{Data Statistics and Analysis}

We have demonstrated the statistics of objects and dialogues in \invig dataset.

\paragraph{Images and Objects} 

As shown in \tabref{tab:data_comp}, \invig dataset totally contains more than 520K images with 43M segmented object instances.
Each image is labeled with a multi-turn goal-oriented disambiguation dialogue.
Among all the data, we have question-answer pairs, more than 4.4M sentences, and around 80M tokens.
To the best of our knowledge, this is the first large-scale dataset for the task of open-ended interactive visual grounding and disambiguation.

\paragraph{Human-to-Human Dialogues}

\invig includes 21K human-to-human dialogues from crowd-sourcing based on 311K object instances belonging to 510 different categories.
The distribution of dialogue rounds and sentence lengths are shown in \figref{fig:invig21k_dialog_stat}.
We can see that more than 99\% dialogues end in 6 interaction rounds, and simultaneously, more than 80\% sentences include less than 10 words.
We also demonstrate the word clouds of the dialogues in \figref{fig:invig21k_word_cloud}.
The first column shows the overall word clouds for questions and answers, and the following ones illustrate word distributions by turns.
We can see that people used to ask questions like ``which one'' in the first turn, and then move on to confirming questions.
The end signals (e.g. ``thank'') usually appear after the 3rd turn of dialogues.
Moreover, interestingly, the ratio of spatial words decreases gradually with an increasing number of dialog turns, meaning that spatial relationships are preferred to answer the question ``which one''.
These observations accord with our experience, which might be considered in the design of interaction models.



      

\paragraph{Automaticlly Generated Dialogues}

We illustrate the distribution of automatically generated dialogues in \figref{fig:invig500k_dialog_stat}.
We can see that after the rephrase of ChatGPT, dialogues in \invig 500K are much more complex than human utterances.
The maximum number of dialogue turns increases to more than 20 and the maximum length of each sentence is larger than 100 words.
We also visualize the word clouds for \invig 500K in \figref{fig:invig500k_word_cloud}.
Similar to \invig 21K, spatial relationships are used as answers mostly in the 1st turn while the confirming answers like ``Yes'' mostly appear after the 2nd dialogue turn.


\begin{table*}[t]  
\caption{M-VQA Performance}  
\begin{tabularx}{\textwidth}{>{\raggedright\arraybackslash}p{98pt}*{10}{>{\centering\arraybackslash}X}}   
\toprule  
& BLEU-1 & BLEU-2 & BLEU-3 & BLEU-4 & METEOR & ROUGE & CIDEr   & R@1 & R@5 & Rank \\ 

\midrule  
  XVLM 21K & 24.6  & 15.8 & 10.4 & 7.2 & 16.0 & 28.3 & 72.8 & 49.8 & 73.9 & 5.0 \\  
  \invigmodel 21K & 26.7 & 17.8 & 12.1 & 8.6 & 16.9 & 30.2 & 81.9 & \bf 50.9 & \bf 75.0 & \bf  4.8\\  
  \invigmodel 500K & 23.2 & 14.5 & 9.5 & 6.7 & 15.4 & 22.8 & 52.3 & 39.7 & 61.5 & 7.1\\  
  \invigmodel 500+21K & \bf 27.7 & \bf 19.3 & \bf 14.0 & \bf 10.5 & \bf 17.8 & \bf 33.4 & \bf 96.2 & 49.9 & 74.3 & \bf 4.8\\ 
\bottomrule  
\end{tabularx}  
\label{tab:vqa}
\end{table*}  

\begin{table*}[t]  
\caption{M-VQG Performance}  
\begin{tabularx}{\textwidth}{>{\raggedright\arraybackslash}p{98pt}*{10}{>{\centering\arraybackslash}X}}   
\toprule  
& BLEU-1 & BLEU-2 & BLEU-3 & BLEU-4 & METEOR & ROUGE & CIDEr   & R@1 & R@5 & Rank \\ 

\midrule  
  \invigmodel-MT 21K & 31.0 & 24.9 & 21.1 & 18.3 & 16.2 & 48.2 & 117.4 & 38.0 & 63.6 & 5.8 \\  
  \invigmodel-BBOX 21K & 34.2 & 27.1 & 22.8 & 19.5 & 17.3 & 49.2 & 120.5 & 49.9 & 68.2 & 5.4\\  
  \invigmodel 21K & \bf 34.7 & 27.6 & 23.1 & 19.9 & 17.4 & 49.7 & 122.2 & 49.9 & 67.9 & 5.4 \\  
  \invigmodel 500K & 24.0 &15.8 & 10.9 & 7.9 & 13.9 & 38.7 & 36.8 & 33.1 & 58.1 & 6.9 \\  
  \invigmodel 500+21K & 34.2 & \bf 27.7 &\bf  23.5 & \bf 20.4 & \bf 17.6 & \bf 50.3 & \bf 128.2 & \bf 52.6 & \bf 74.7 & \bf 4.5 \\ 
\bottomrule  
\end{tabularx}  
\label{tab:vqg}
 \vspace{-10pt}
\end{table*}  

\section{BASELINES AND EXPERIMENTAL RESULTS}

Based on the collected human-to-human dialogues, we conduct a series of downstream tasks to set up a benchmark for interactive visual grounding.
We aim to answer the following questions in this section:

(1) What tasks can we do by taking advantage of \invig dataset? (\secref{sec:tasks})

(2) How can \invig 21K and \invig 500K be used to perform better in disambiguation tasks? (\secref{sec:results})

(3) What are the pros and cons of state-of-the-art visual-language models for disambiguation tasks? (\secref{sec:results})

\subsection{Tasks}
\label{sec:tasks}

We design four downstream tasks based on our \invig dataset: a) Multi-Turn Visual Question Answering; b) Visual Question Generation; c) Multi-Turn Visual Grounding; d) Interactive Visual Grounding.

\paragraph{Multi-turn Visual Question-Answering (M-VQA)} requires the model to answer open-ended questions based on image observations and dialogue history, hence, the inputs are an image $\image$, an initial referring expression $\expr$, and a group of questions and answers $\{(\q{i},\ans{i})\}_{i=1}^{T-1}$, and the question to be answered $\q{T}$.
The output is a sequence of words that form an answer $\ans{T}$.

\paragraph{Multi-turn Visual Question Generation (M-VQG)} requires the model to generate a question for clarification in the next turn. Therefore, the input includes an image $\image$, an initial referring expression $\expr$, and a group of questions and answers $\{(\q{i},\ans{i})\}_{i=1}^{T}$.
The output should be a sequence of words that form a question of the next turn.

\paragraph{Multi-turn Visual Grounding (M-VG)} is defined as grounding the target object $\tgt=(x_{min}, y_{min}, x_{max}, y_{max})$ given an image $\image$, an initial referring expression $\expr$, and a dialogue history $\{(\q{i},\ans{i})\}_{i=1}^{T}$ that contains the complete information about the targets.

\paragraph{Interactive Visual Grounding (I-VG)} is a new task that involves both active interaction and interactive visual grounding.
In this setting, the ground truth dialogue history is not available.
Instead, the agent needs to actively interact with humans to collect the necessary information for the final grounding task.
In previous works, humans are usually involved in evaluating the performance of a model \cite{shridhar2020ingress}, which is time-consuming and also expensive to evaluate a large dataset like our \invig.
Instead, we introduce an Oracle by following \cite{de2017guesswhat} in this paper to conduct the automatic evaluation.
The Oracle follows the same form as the multi-turn visual question-answering model.

Formally, the input in this task contains an image $\image$ and an initial referring expression $\expr$.
The model needs to generate a set of questions $\{\q{i}\}_{i=1}^{T}$ to finally ground the target object $\tgt=(x_{min}, y_{min}, x_{max}, y_{max})$.
Meanwhile, the Oracle model should play the role of users, and give answers $\{\ans{i}\}_{i=1}^{T}$ to the corresponding questions.
High performance in this task requires seamless coordination of all the above three tasks.

\subsection{Metrics}

\paragraph{M-VQA and M-VQG}
Since we do not restrict the question and answer space, we introduce two sets of metrics for the evaluation of these two tasks.
Firstly, we follow previous works in natural language processing and measure the similarity between the predictions and the ground truths.
In detail, we use BLEU-1 and BLEU-4 \cite{papineni2002bleu}, CIDEr \cite{vedantam2015cider}, METEOR \cite{banerjee2005meteor}, and ROUGE \cite{lin2004rouge} as the metrics for text similarity. 
Nevertheless, these metrics assume that the generated text should be unique, which violates the intuition of open-ended dialogues.
Hence, we also introduce the retrieval-based multi-choice metrics: \textit{Recall@k} and \textit{GT Rank}.
To do so, we assign 30 candidates for each question and answer.
To make the choices challenging, the candidates are sampled from the dialogues of the top 30 similar images using CLIP \cite{radford2021learning} embeddings.
The model is required to rank all candidates together with the ground truth.
\textit{Recall@k} measures the fraction of the ground truth being in top-$k$ candidates.
\textit{GT Rank} measures the mean rank of the ground truth among all candidates.

\paragraph{M-VG and I-VG}

We follow the traditional settings \cite{yu2016modeling} and use accuracy to evaluate the performance of Interactive Visual Grounding.
Concretely, for each model, we measure the fraction of predicted bounding boxes that have the Interaction of Union (IoU) no smaller than $m$, given the dialog histories, where $m \in \{0.1, 0.2, ..., 1.0\}$.

\subsection{Baselines}

\begin{figure}[t]
 \center{\includegraphics[width=0.5\textwidth]{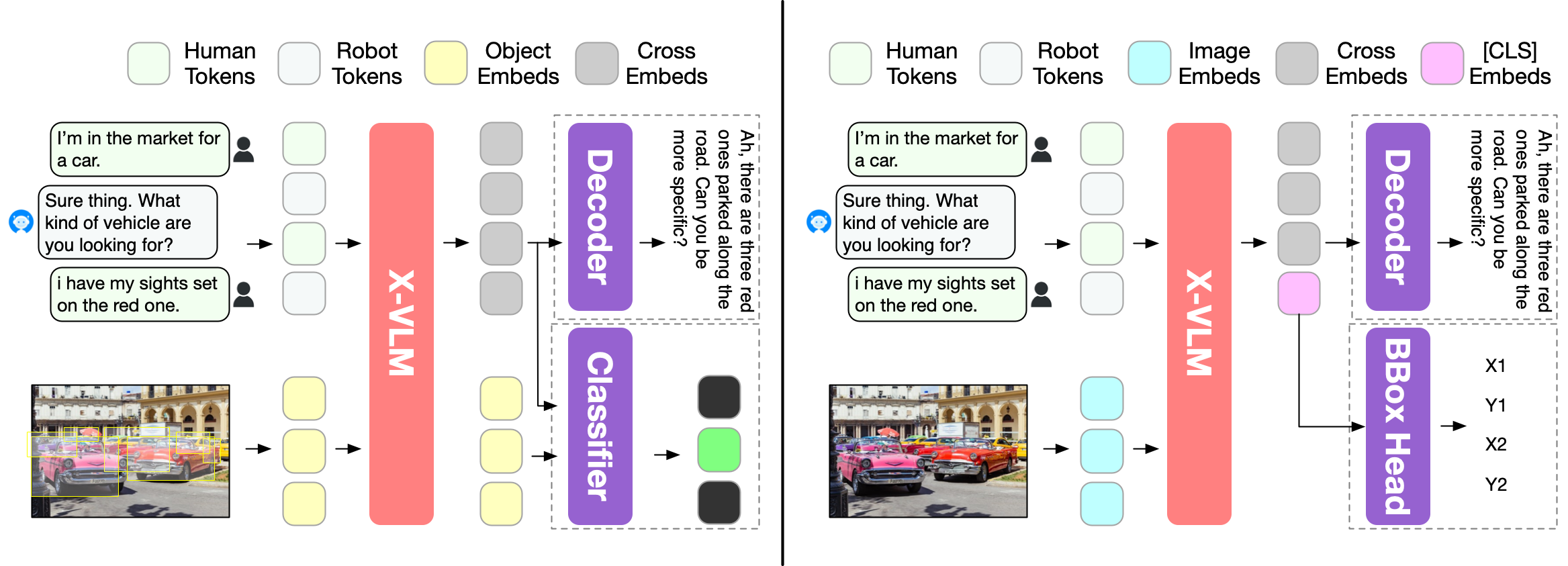}}
 \caption{Networks of baseline methods. Left: \invigmodel. Right: \invigmodel BBOX. Note that \invigmodel BBOX is conditioned on image embeddings and does not rely on an external bounding box detector.}
 \label{fig:invig_vlm_arch}
     \vspace{-10pt}
\end{figure}

We implement a simple yet efficient baseline algorithm named \invigmodel, based on the state-of-the-art vision-language foundation model X-VLM \cite{zeng2021multi}. 
It includes three parts: an \textit{Oracle} model which can answer open-ended questions, a \textit{Questioner} model that interacts with the Oracle to collect information, and a 
\textit{Guesser} model to guess targets with the input of ground truth bounding boxes of all objects or the ones detected using Detic \cite{zhou2022detecting}.
The overall architecture is shown in \figref{fig:invig_vlm_arch}.
To validate different formulations for multi-turn dialogues, we also implement a multi-turn variant called \invigmodel-MT, following previous work \cite{lu2019vilbert}.
It re-formulates the Agent in a multi-turn mode.
In each turn, the model takes as inputs the belief of each object from the last turn as well as the current textual and visual embeddings.
The third one is similar to \invigmodel, which is also demonstrated in \figref{fig:invig_vlm_arch}.
In this version, the visual grounding head is replaced with a bounding box regressor rather than a classifier following the formulation from \cite{kamath2021mdetr, zeng2021multi}.
It gets rid of object proposals and directly outputs the location of the target.
For M-VQA, we also compare to the original X-VLM with only the question to be asked as language inputs, which is trained on our \invig 21K dataset (XVLM 21K).

\subsection{Results}
\label{sec:results}

We do experiments for all four tasks introduced in \secref{sec:tasks} on \invig 21K and 500K datasets.

\begin{figure}  
    \centering  
      
    \begin{subfigure}[b]{0.31\textwidth}  
        \centering  
        \includegraphics[width=\textwidth]{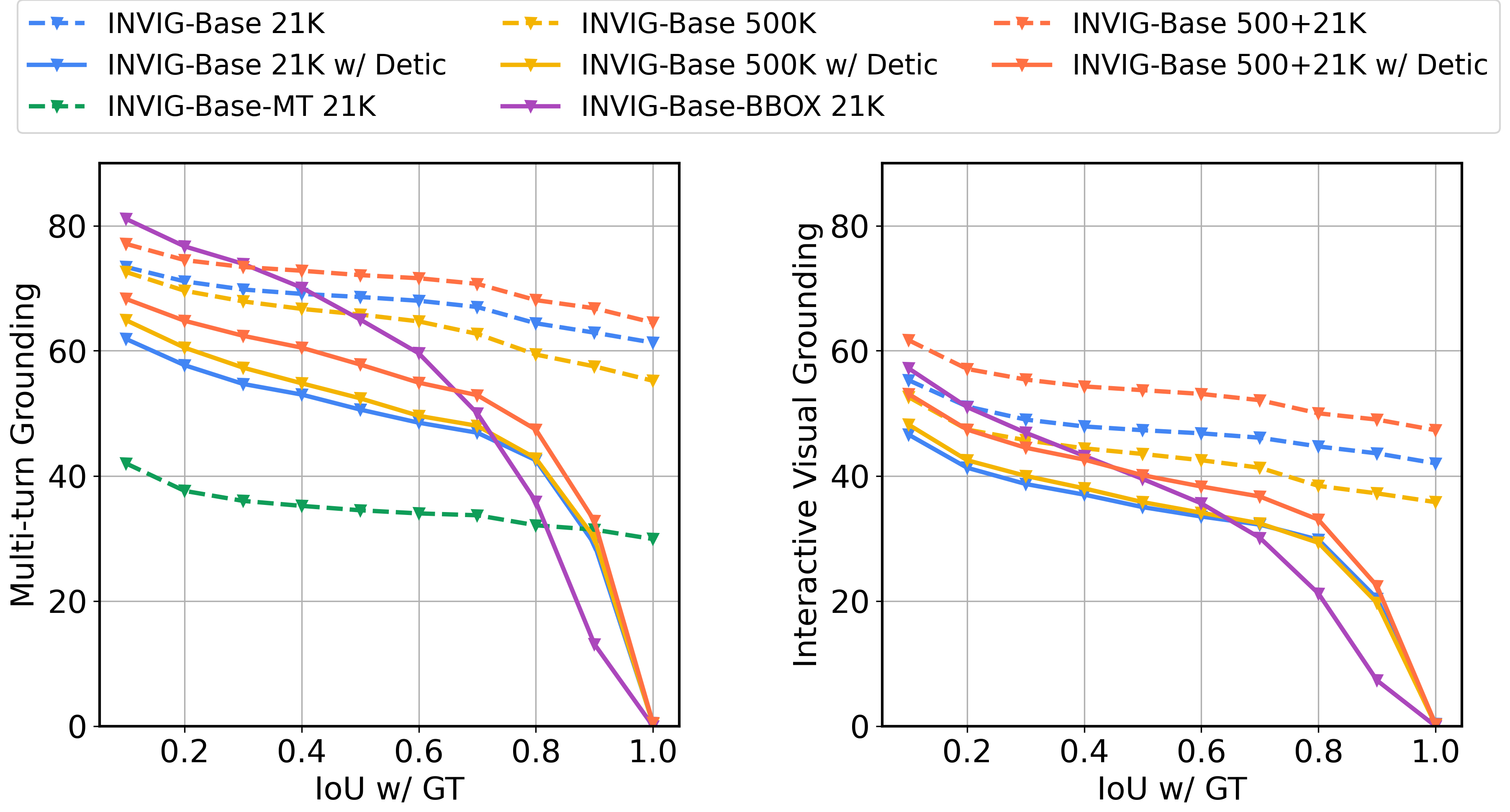}  
        \caption{Left: M-VG. Right: I-VG}  
        \label{fig:M-VG_I-VG_res}  
    \end{subfigure}  
    \hfill  
    \begin{subfigure}[b]{0.155\textwidth}  
        \centering  
        \includegraphics[width=\textwidth]{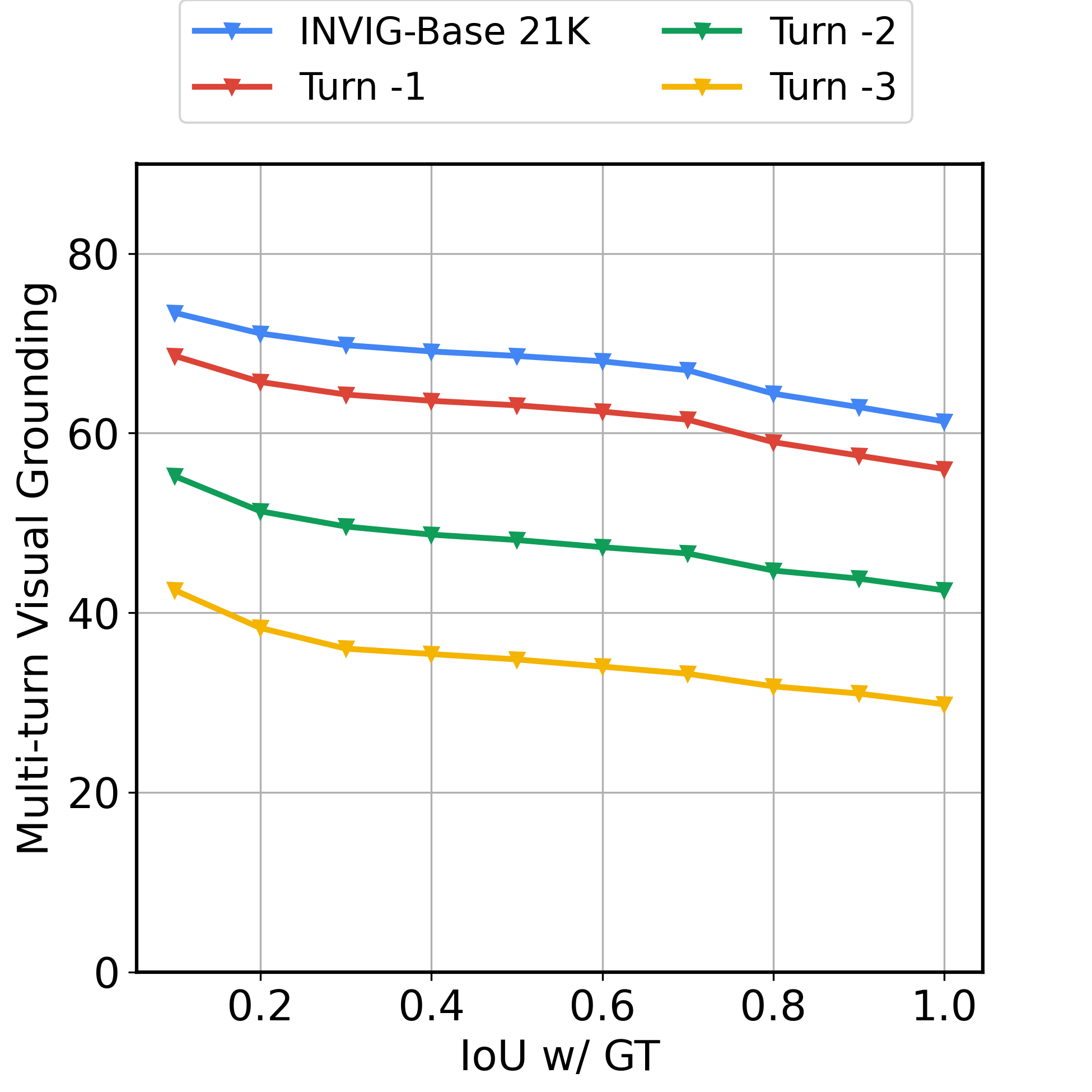}  
        \caption{}  
        \label{fig:g_less_turns}  
    \end{subfigure}  
      
    \caption{Multi-Turn Visual Grounding (M-VG) and Interactive Visual Grounding (I-VG) performance.}  
    \label{fig:M-VG_I-VG_all_res}  
    \vspace{-10pt}
\end{figure}

\paragraph{Pre-training improves performance}
\label{sec:pretraining}
For the text generation tasks, specifically M-VQA and M-VQG (see Table \ref{tab:vqa} and Table \ref{tab:vqg}), training on a large amount of noisy data (\invigmodel 500K) yields inferior performance compared to training on a smaller dataset with less clean data (\invigmodel 21K). 
However, pre-training on a large amount of noisy data still leads to a significant improvement in performance. 
Notably, the pre-trained model \invigmodel 500+21K outperforms the model without pre-training across most metrics.
For M-VG and I-VG (\figref{fig:M-VG_I-VG_res}), we also have similar conclusions.

\paragraph{Dialog history is crucial}
\label{sec:dlg_his}
In the case of M-VG, as shown in \figref{fig:g_less_turns}, \textit{Turn-}$m$, $m\in \{1,2,3\}$, means grounding performance without the last $m$ dialogue turns.
We can see that the grounding performance drops consistently with less information.
Besides, dialog history also helps M-VQA.
As shown in \tabref{tab:vqa}, in the case of M-VQA, our \invigmodel consistently outperforms the original X-VLM, albeit by a small margin. 
This observation highlights two key findings: 
1) Dialog histories contribute to enhanced performance in M-VQA, and 
2) Most questions can be reasonably answered without relying on dialog history.

\paragraph{Target classification performs stably}
\label{sec:diff_form}
From \figref{fig:M-VG_I-VG_res}, we can conclude that when considering different visual grounding approaches, direct bounding box prediction demonstrates superior performance at lower IoU thresholds, but its performance declines significantly as the threshold increases.
In contrast, the performance of \invigmodel based on the target classification is more stable.
We can see that with ground truth object bounding boxes, \invigmodel 500+21K still achieves more than 60\% success rate with ground truth dialogues.
When paired with Detic, its performance is stable when the threshold increases.
Besides, \invigmodel-MT exhibits the poorest performance, suggesting that the multi-turn dialogue formulation is not optimal for interactive visual grounding tasks.

\section{CONCLUSIONS}

In this paper, we present \invig dataset, the first large-scale dataset including more than 520K images and dialogues for interactive visual grounding, to resolve the challenge of language ambiguity in HRI.
We have conducted extensive and comprehensive experiments and set up a suite of baseline solutions to resolve HRI ambiguity.
Our results demonstrate that based on the \invig dataset, the robot can successfully disambiguate interaction with a success rate of 45.6\% in challenging and realistic scenarios in validation.
Future works include developing an automatic data-cleaning process for high-quality data collection, integration into more interactive robot systems, and validation of performance on more downstream interactive robot tasks.

    
    
    








\bibliographystyle{unsrt}
\bibliography{ref}

\vspace{12pt}

\end{document}